\documentclass[10pt,journal,compsoc]{IEEEtran}
\usepackage{graphicx}
\usepackage{amsfonts}
\usepackage{amsmath}
\pdfoutput=1 
\usepackage{booktabs}
\usepackage{xcolor}
\usepackage{color}
\usepackage{multirow}
\usepackage[normalem]{ulem}
\usepackage{float}
\useunder{\uline}{\ul}{}
\begin{document}

\title{Spatial-Temporal Interactive Dynamic Graph Convolution Network for Trafﬁc Forecasting}

\author{Aoyu Liu, Yaying Zhang
\IEEEcompsocitemizethanks{\IEEEcompsocthanksitem A. Liu and Y. Zhang are with the Key Laboratory of Embedded System and Service Computing, Ministry of Education, Tongji University, Shanghai 201804, China. (email: \{liuaoyu, yaying.zhang\}@tongji.edu.cn)}
\thanks{Manuscript received xx xx, 202x; revised xx xx, 202x.}}

\markboth{Journal of \LaTeX\ Class Files,~Vol.~xx, No.~x, xxx~202x}%
{Shell \MakeLowercase{it{et al.}}: Bare Demo of IEEEtran.cls for Computer Society Journals}
\IEEEtitleabstractindextext{%
\begin{abstract}
Accurate traffic forecasting is essential for urban traffic control, route planning, and flow detection. Although many spatial-temporal methods are currently proposed, they are still deficient in synchronously capturing the spatial-temporal dependence of traffic data. In addition, most methods ignore the hidden dynamic associations that arise between the road network nodes as it evolves over time. We propose a neural network-based Spatial-Temporal Interactive Dynamic Graph Convolutional Network (STIDGCN) to address the above challenges for traffic forecasting. Specifically, we propose an interactive dynamic graph convolution structure which divides the traffic data by intervals and synchronously captures the divided traffic data‘s spatial-temporal dependence through an interactive learning strategy. The interactive learning strategy motivates STIDGCN effective for long-range forecasting. We also propose a dynamic graph convolution module through a novel dynamic graph generation method to capture the dynamically changing spatial correlations in the traffic network. Based on a priori knowledge and input data, the dynamic graph generation method can generate a dynamic graph structure, which allows exploring the unseen node connections in the road network and simulating the dynamic associations between nodes over time. Extensive experiments on four real-world traffic flow datasets demonstrate that STIDGCN outperforms the state-of-the-art baselines.
\end{abstract}

}

\maketitle

\IEEEdisplaynontitleabstractindextext

\IEEEpeerreviewmaketitle

\IEEEraisesectionheading{\section{Introduction}\label{sec:introduction}}
\IEEEPARstart{W}ith the help of available massive urban traffic data collected from  sensors on the road, cabs, private car trajectories, and transaction records of public transportation, big traffic data analysis has become an indispensable part of smart city development \cite{ITS} for traffic planning, control, and condition assessment. Traffic forecasting, aiming to predict the urban dynamics with the observed historical traffic data, is critical for traffic services like flow control, route planning, and flow detection. Accurate traffic forecasting can contribute to reducing road congestion, facilitating city management of the traffic road network, and even enhancing transportation efficiency  \cite{survey}.

Although traffic forecasting has been continuously an active research hot-spot in the past several decades and extensive research efforts have been made in this field to improve predictive performance, it still faces some challenges. The traffic data is spatial-temporal data with complex temporal correlations and dynamic spatial correlations. As a type of time series data, urban traffic data shows specific periodicity and trends, such as morning and evening peaks. Effective capture of periodicity and trend requires models that can accurately capture the long-term dependence between output and input. These complex temporal correlations hence make long-range forecasting of traffic data difficult. For example, when using the past 12-time steps’ observed traffic data to predict the future 12-time steps’ data, it is generally more difficult to accurately predict the 9th-12th time steps’ data than to predict 1st-3rd time steps’ data. As shown in Figure 1, traffic data’s dynamic spatial correlations are also diverse due to the intricate traffic flow on the road network. Figure 1a shows that traffic conditions can spatially affect each other and change dynamically. For example, an accident on one road segment can affect the traffic conditions of its nearby road segments. Besides, traffic flow in different directions on the same segment may also behave differently.

\begin{figure}[htbp]  
\centering
\includegraphics[width=\linewidth]{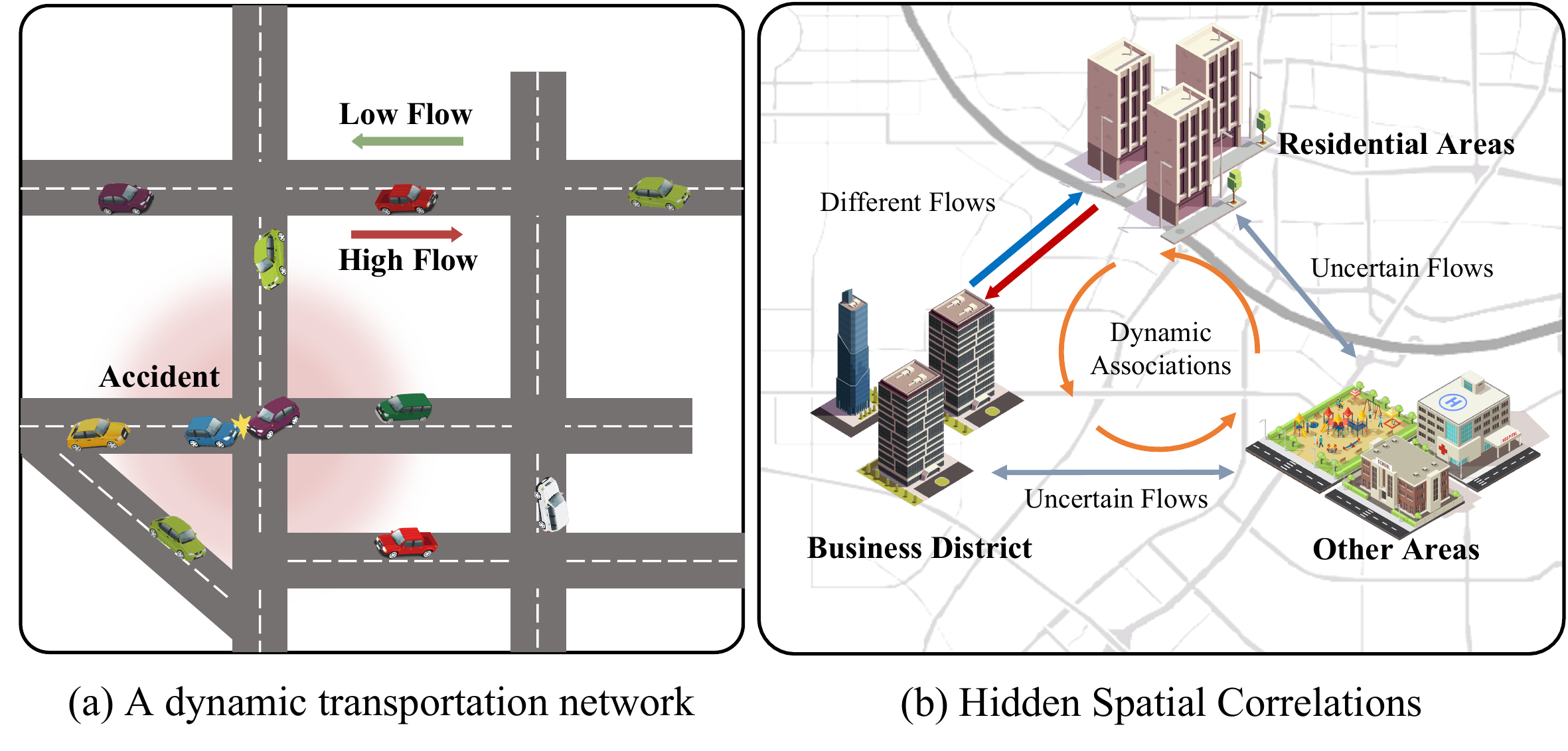}
\caption{Spatial diversity of traffic data. Traffic flow in (a) shows that traffic flow in different directions on the same road may behave differently, and accidents on the road can affect nearby roads. (b) shows the hidden spatial correlation(spatial heterogeneity, dynamic associations, and uncertainty).}      
\label{fig1}   
\end{figure}

\begin{table*}[h]
\caption{Classification of spatial-temporal modeling methods.}
\centering
\resizebox{17cm}{!}{
\begin{tabular}{c|c|c|c}
\toprule
Model         & Modeling Construction & Spatial Correlation               & Temporal Correlation \\ \hline
STGCN\cite{STGCN}         & In Series        & GCN (Pre-Defined Graph)           & GLU                 \\
ST-MetaNet\cite{ST-MetaNet}    & In Series        & GAT (Pre-Defined Graph)           & GRU                 \\
ASTGCN\cite{ASTGCN}        & In Series        & Attention+GCN (Pre-Defined Graph) & 1D CNN+Attention    \\
Graph WaveNet\cite{GraphWaveNet} & In Series        & GCN (Adaptive Graph)              & TCN                 \\
MGDCN\cite{MGDCN}         & In Series        & GCN (Multiple Graph)              & GRU+Attention       \\
SLCNN\cite{SLCNN}         & In Series        & GCN (Dynamic Graph)              & 1D CNN              \\
MTGNN\cite{MTGNN}         & In Series        & GCN (Adaptive Graph)              & 1D CNN              \\
STGODE\cite{STGODE}        & In Series        & ODE+GCN (Dynamic Graph)           & TCN                 \\
DMSTGCN\cite{DMSTGCN}       & In Series        & GCN (Adaptive Graph)              & 1D CNN              \\
ASTGNN\cite{ASTGNN}        & In Series        & GCN (Dynamic Graph)               & Attention           \\
STG-NCDE\cite{STG-NCDE}      & In Series        & NCDE+GCN (Adaptive Graph)         & NCDE                \\ \hline
GMAN\cite{GMAN}          & In Parallel      & Attention (Pre-Defined Graph)     & Attention           \\
STFGNN\cite{STFGNN}        & In Parallel      & GCN (Multiple Graph)              & 1D CNN              \\ \hline
DCRNN\cite{DCRNN}         & In Embedded      & GCN (Pre-Defined Graph)           & LSTM                \\
AGCRN\cite{AGCRN}         & In Embedded      & GCN (Adaptive Graph)              & GRU                 \\
MRA-BGCN\cite{MRA-BGCN}      & In Embedded      & Attention+GCN (Multiple Graph)    & GRU                 \\
STSGCN\cite{STSGCN}        & In Embedded      & GCN (Multiple Graph)              & GCN(Sliding Window)\\
\toprule	
\end{tabular}
}
\end{table*}

There are also some hidden spatial correlations in the traffic dynamics, which stem from spatial heterogeneity, dynamic associations, and uncertainty. This implicit spatial correlation also poses a challenge for traffic forecasting. As in Figure 1b, spatial heterogeneity means that different areas(e.g., residential areas and business districts) have different traffic patterns because they have different characteristics, such as road types, road width, POIs, etc. Dynamic associations are time-varying associations between nodes generated by traffic flows. This association could be learned from historical traffic data and road network structure. Uncertainties refer to the impact of events on traffic conditions, such as weather changes, holidays, and emergencies.

To effectively capture the spatial-temporal dependence, some deep learning-based methods are applied. The spatial dependence and temporal dependence are widely studied in traffic forecasting. As shown in Table 1, some methods captured the temporal correlation and spatial correlation separately and combined them either in serial or in parallel. These methods might weaken the captured spatial-temporal correlation and even amplify some unimportant features. Some other recent methods try to synchronously capture the spatial and temporal dependencies by embedding the spatial module into the temporal module, which contributes positively to the models. Nevertheless, the spatial-temporal modules used in these methods are limited in interactive learning of the temporal and spatial information between traffic data, which affects the model's awareness of the periodicity and trend of the sequence.

To capture the hidden spatial correlations, many current studies map out deeper graph structures by defining various adjacency matrices. As shown in Table 1, these studies capture the hidden spatial correlations by defining adaptive adjacency matrices or combining multiple adjacency matrices. However, these methods do not adequately utilize the historical traffic data. Although the adaptive adjacency matrix could discover the implicit relationship between graph nodes and thus enhance the model's capture of spatial heterogeneity, it would be fixed as the model training stops. It could not simulate the dynamic association between graph nodes over time. Therefore, these methods are still insufficient to capture the hidden spatial correlations effectively. The dynamic association between nodes over time could be further explored with historical traffic data and initial adjacency matrix.

With the above concerns, we propose a Spatial-Temporal Interactive Dynamic Graph Convolutional Neural Network (STIDGCN) to explore the interactions between input data and the dynamic correlations in the road network. We design an interactive dynamic graph convolution network to capture the traffic data's spatial-temporal dependence. By embedding the graph convolution module into an interactive learning structure, STIDGCN can synchronously capture the temporal and spatial correlation of traffic data. This structure exploits the trendiness of time series to divide the series at intervals. The interaction learning between the divided sub-series explores the potential associations between spatial-temporal data. In order to effectively capture the hidden spatial correlations, we propose a dynamic graph convolution network that can fully use existing a priori knowledge (graph structure, historical data) through a dynamic graph generation method.

The contribution of this work is summarized as follows:
\begin{itemize}
	\item A new spatial-temporal model STIDGCN is proposed, which embeds the graph convolution into an interactive learning structure. It enables the synchronous capture of temporal and spatial correlations. Spatial-temporal dependence can be learned through the interactive learning structure and dynamic graph convolution network, thus enabling effective long-range prediction.
	\item We propose a dynamic graph convolution network through a dynamic graph generation method. The dynamic graph is generated by fusing the adaptive and learnable adjacency matrix in which the adaptive adjacency discovers the spatial heterogeneity and learnable adjacency matrix simulates the dynamic associations between nodes.	
	\item Extensive experiments were conducted on four real-world datasets from previous work. The experimental results show that our model has state-of-the-art performance compared to the baseline models.
\end{itemize}

The rest of this paper is organized as follows: Problem definition and related works are stated in Section~\ref{Problem Definition and Related Work}. The proposed method is elaborated in Section~\ref{Spatial-Temporal Interactive Dynamic Graph Convolution Network}. Experimental results and analysis are presented in Section~\ref{Experiment Results and Analysis}. Finally, the conclusion is made in Section~\ref{Conclusion}. 

\section{Problem Definition and Related Work} \label{Problem Definition and Related Work}
\subsection{Problem Definition}
Traffic forecasting is a time series prediction task using a priori knowledge. Here the priori knowledge is the initial adjacency matrix generated from the traffic road network. Given a road network, we can represent it as a graph $G=(V,E,A)$, where $V$ denotes the set of nodes,  $|V|=N$. Each node denotes an observation point (sensor or road segment) in the traffic road network, and $E$ is a set of edges. The edges in graph $G$ represent the connection relationships between the nodes, and the weights of the edges are the distances between the nodes. $A \in \mathbb{R}^{N \times N}$ is the adjacency matrix of graph G. The traffic forecasting task is to predict the future traffic sequence $X_G^{(t+1)},X_G^{(t+2)},\ldots,X_G^{(t+T^\prime)}$ using a segment of historical sequence $X_G^{(t-T+1)},X_G^{(t-T+2)},\ldots,X_G^{(t)}$, where $X_G^{(t)}\in\mathbb{R}^{N\times C}$ denotes the observation of graph $G$ at time step $t$, $C$ denotes the number of feature channels, $T$ denotes the length of a given historical time series, $T^\prime$ denotes the length of the time series to be predicted.The final traffic flow forecasting problem can be defined as:
\begin{equation}
\left[X_{G}^{\left(t-T+1\right)}, \ldots, X_{G}^{(t)}\right] \stackrel{f}{\rightarrow}\left[X_{G}^{(t+1)},\ldots, X_{G}^{\left(t+T^{\prime}\right)}\right]
\end{equation}where $f$ denotes the learning function from the historical sequence to the predicted sequence.

\subsection{Spatial-Temporal Trafﬁc Forecasting}
Traffic forecasting has been extensively studied and traffic forecasting methods can be divided into two groups: traditional methods and deep learning methods. The traditional methods include classical statistical methods and traditional machine learning methods. 

Early statistical methods used for traffic forecasting are represented by models such as Historical Averages (HA), Auto-Regressive Integrated Moving Average (ARIMA) \cite{ARIMA}, and Vector Auto-Regressive (VAR) \cite{VAR}. These models are based on linear time series methods and need to rely on static assumptions. Since traffic data are complex nonlinear data, these models naturally underperform compared to machine learning-based methods.

To capture complex nonlinear relationships in traffic data, some traditional machine learning methods are applied to traffic forecasting, such as Support Vector Regression (SVR) \cite{SVR}, Random Forest Regression (RFR) \cite{RFR}, and K-Nearest Neighbor (KNN) \cite{KNN}. These methods tend to be more effective but require certain experience to design manual features.
\begin{figure*}[htbp]  
\centering
\includegraphics[width=\linewidth]{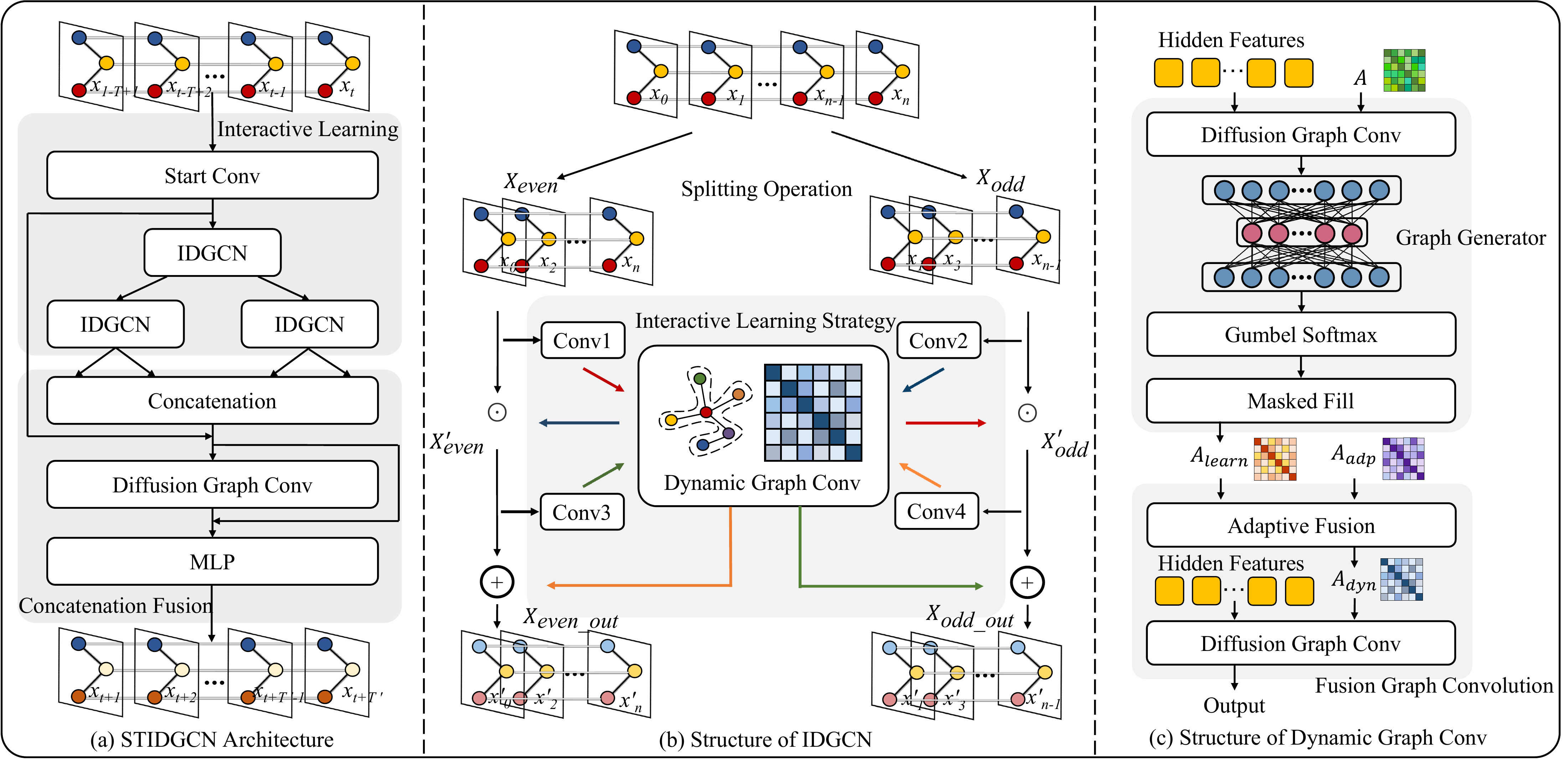}
\caption{STIDGCN architecture. STIDGCN consists of an interactive learning module and concatenation fusion module. The interactive learning module uses a tree structure and divides the input data at intervals. The dynamic graph convolution module is embedded into the IDGCN to achieve synchronous extraction of spatial-temporal dependencies. DGCN consists of a graph generator and fusion graph convolution. Graph generator generates a learnable adjacency matrix to simulate the dynamic association of changes between nodes. The fusion graph convolution fuses two adjacency matrices and performs graph convolution.} 
\label{fig2}   
\end{figure*}

Deep learning based methods are effective in automatically capturing features for representational learning. Early studies used RNNs \cite{LSTM}, \cite{GRU}, \cite{GRU/}, \cite{Bi-LSTM} (including LSTM and GRU) to capture the temporal features of traffic data. Ma et al. \cite{LSTM} first applied LSTM to traffic speed forecasting. LSTM improves the ability to model long-range sequential correlations while overcoming the gradient disappearance and explosion problems. Cui et al. \cite{Bi-LSTM} proposed a Stacked Bidirectional and Unidirectional LSTM for traffic speed forecasting to capture the backward and forward time series correlation. The RNN-based methods have its limitations, such as error accumulation, slow training, and inability to handle long sequences. Convolutional neural network(CNN) can process data in parallel and have a relatively small memory occupation. Some CNN-based methods are widely used for time series, such as WaveNet \cite{WaveNet} and Temporal Convolutional Neural Network (TCN) \cite{TCN}. Recently, Liu et al. proposed SCINet \cite{SCINet} to expand the receptive ﬁeld of convolutional operations and achieve multi-resolution analysis in a downsample-convolve-interact manner. This novel time series forecasting method also obtains outstanding performance in traffic forecasting. However, the above methods do not consider the spatial correlation of traffic data to fully reflect the Spatial-Temporal dependence. 

Some studies introduce CNN to capture spatial correlation of traffic data. Zhang et al. \cite{ST-ResNet} proposed ST-ResNet to predict urban pedestrian flow with a CNN structure. Yao et al. \cite{STDN} proposed a spatial-temporal dynamic network for New York cab and bicycle data forecasting, which processed spatial-temporal information by CNN and LSTM, respectively. CNN treats traffic data as Euclidean data to capture spatial correlations. Practically the traffic road network has its own topology and traffic data is essentially non-Euclidean data, and the spatial correlation captured by CNN based methods is limited. Graph Neural Network(GNN) can model non-Euclidean data, and it is suitable for capturing spatial correlation in spatial-temporal data. 

Recently, GNN based spatial-temporal modeling methods are widely used  to capture the spatial correlation of traffic data. Meanwhile, the RNNs-based, CNNs-based, and attention mechanism-based \cite{attention} modules capture the temporal correlation in spatial-temporal modeling methods. DCRNN \cite{DCRNN} models the spatial correlation of traffic as a diffusion process on directed graphs, and it uses GRU in combination with diffusion GCN for traffic forecasting. STGCN \cite{STGCN} employs convolution operation fully in the time dimension and uses spectral graph convolution to capture the spatial correlation of traffic data. Since urban traffic is a dynamically changing system, the fixed graph structure with a static adjacency matrix cannot represent such dynamics. For this reason, Wu et al. \cite{GraphWaveNet} proposed Graph WaveNet to capture the dynamic spatial correlation by designing an adaptive adjacency matrix with GCN. Song et al. \cite{STSGCN} proposed STSGCN to capture temporal and spatial correlations synchronously by using sliding windows to process the constructed local spatial-temporal graphs. Nevertheless, these approaches have not effectively capture the hidden dynamic associations between nodes, which inherently exists in traffic dynamics.

Besides, methods based on the self-attention mechanism are widely used in the long-range forecasting of traffic data. Guo et al. \cite{ASTGCN} proposed ASTGCN as a spatial-temporal graph convolutional network based on attention mechanisms. The temporal and spatial attention mechanisms were used to model temporal and spatial correlations, respectively. Zheng et al. \cite{GMAN} proposed a Graph Multi-Attention Network (GMAN) with an encoder-decoder structure consisting of spatial-temporal attention blocks to model the effects of spatial-temporal factors on traffic conditions. Guo et al. \cite{ASTGNN} proposed an Attention-based Spatial-Temporal Graph Neural Network (ASTGNN). Their proposed trend-aware self-attention module is used to capture temporal correlation, and the proposed dynamic graph convolution module is used to capture dynamic spatial correlation.

\subsection{Graph Convolution Network}
GCN methods \cite{GNN} extend the traditional convolution method to graph structures and use it to capture the neighbor information of nodes and edges. The two mainstream GCN methods are spectral and spatial methods, respectively. Bruna et al. \cite{GCN} first proposed a generalized GCN based on spectral methods in 2014. They mapped the topological graph structure in the spatial domain to the spectral domain by Fourier transform and performed the convolution operation. Finally, the GCN operation is completed by returning to the spatial domain using the inverse transform. Defferrard et al. \cite{ChebNet} improved the traditional GCN and proposed ChebNet to reduce the complexity of Laplacian computation. Kipf and Welling \cite{ChebNet2} simplified ChebNet and achieved state-of-the-art performance. GCNs based on spatial methods capture the representations of nodes by aggregating feature information from their neighbors for convolution, such as GraphSAGE \cite{GraphSAGE}. In addition, Velikovi and Petar \cite{GAT} proposed GAT, which uses the attention mechanism to dynamically adjust the correlation weights of neighboring nodes and use these weights to determine the node importance.

In contrast to previous work, the proposed STIDGCN shares the spatial-temporal features learned between sequences by using a dynamic graph interactive learning strategy.

\section{Spatial-Temporal Interactive Dynamic Graph Convolution Network}\label{Spatial-Temporal Interactive Dynamic Graph Convolution Network}

We proposed the spatial-temporal interactive dynamic graph convolutional network(STIDGCN) to synchronously capture of the temporal and spatial correlations. In this approach, dynamic graph structures is presented to capture hidden spatial associations and the dynamic associations between nodes over time are explored. The framework is shown in Figure 2a, which consists of two main modules, namely the interactive learning module and the concatenation fusion module.

\begin{figure*}[htbp]  
\centering
\includegraphics[width=12cm]{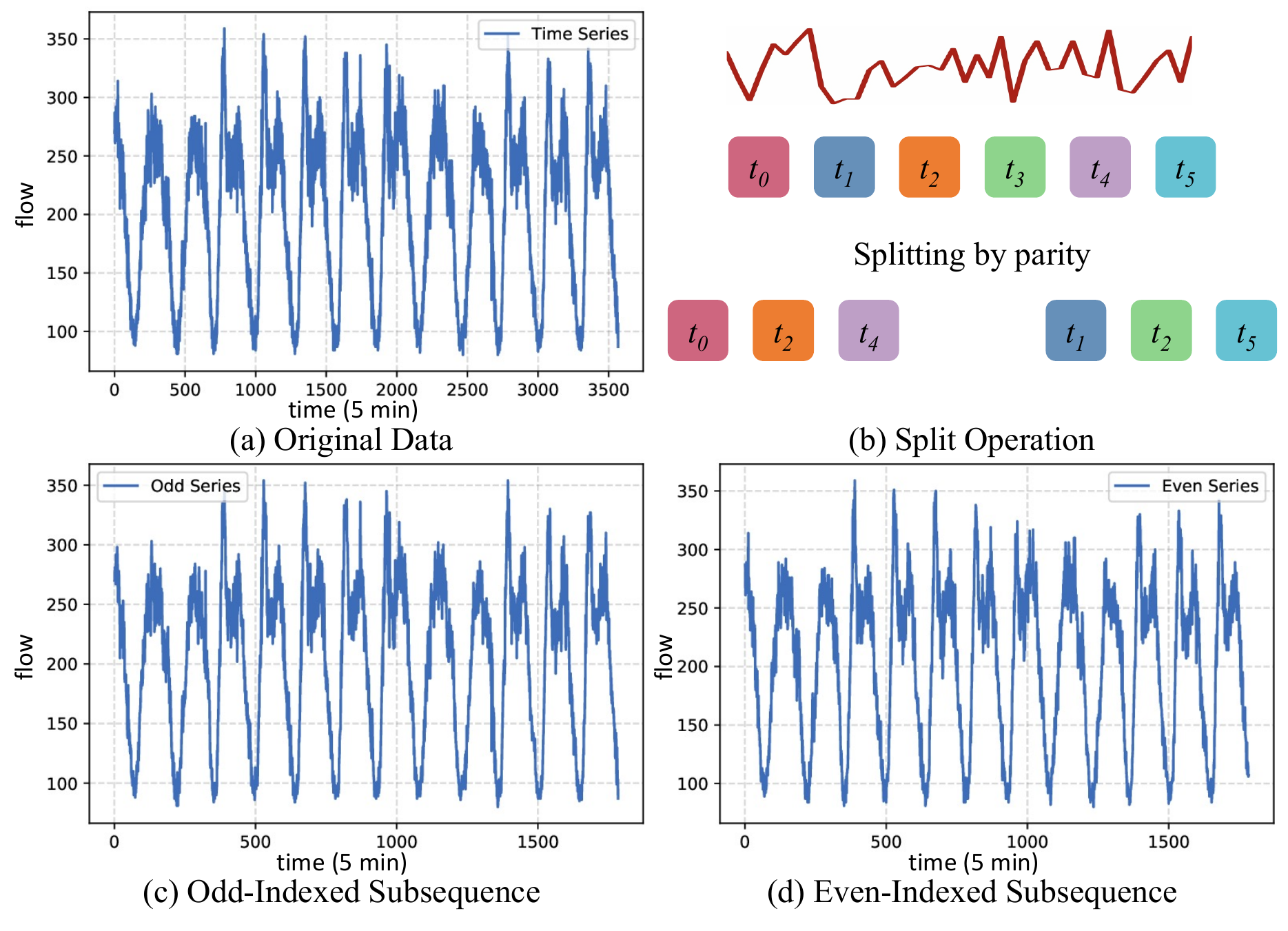} 
\caption{Impact of splitting operations on traffic data. (a) is the initial sequence, (c) is the subsequence generated by dividing by odd index, and (d) is the subsequence generated by dividing by even index. The subseries obtained after the interval division operation still retain the original series's time trend information.} 
\label{fig3}   
\end{figure*}

First, the original input data is passed through a convolutional layer to get a higher dimensional space to capture deeper dependencies. The features are then fed into the interactive Dynamic Graph Convolution Network (IDGCN) module. Here we deal with interactive learning by divide and conquer. As shown in Figure 2b, the input sequence is equally divided into two subsequences using interleaved sampling. Subsequently, the two half-sized subsequences learn interactively in the interactive learning structure, sharing to learn their respective features. While capturing temporal correlations, the Dynamic Graph Convolution Network (DGCN) module is embedded in the interactive learning structure to interactively learn their respective spatial correlations. After IDGCN, two sequences are output. Subsequences (length halved) are recursively generated in the dividing stage. All the result subsequences in the conquer stage are recombined in a time-indexed order and fed into the dynamic graph convolution module for global spatial-temporal feature capture. Finally, the captured features are passed through the Multilayer Perceptron (MLP) to output the forecasting sequence.

\subsection{Interactive Learning}

Inspired by SCINet \cite{SCINet} and Multilevel Wavelet Decomposition Network \cite{wavelet}, sequence data can be downsampled to achieve multi-resolution analysis and expand the receptive ﬁeld. This paper employs CNN and GCN to implement the interactive learning strategy. Compared to CNN-based TCN, the interaction between sequences is tighter, which could better capture the complex temporal correlations and dynamic spatial correlations.

As shown in Figure 2b, the main component of interactive learning is IDGCN. For most time series data, due to the trending and closeness, the interleaved down-sampling subsequences still maintain most information from the original series. As shown in Figure 3, the two subsequences of traffic data for a road segment still preserve the temporal trend and periodicity of the original sequence after downsampling.

The segmented subsequences learn the spatial-temporal features interactively in the dynamic graph convolution module. As shown in Figure 2b, the two subsequences use separate convolution modules to preprocess the features to increase the receptive field. The two subsequences share the parameter weights in the dynamic graph convolution module. More interactive learning processes could be overlayed when necessary.

Assume that $X \in \mathbb{R}^{ C \times N \times T}$ is the input to the interactive learning structure. The subsequence of $X$ after interval division (according to parity index interval) can be represented as $X_{ {odd }} \in \mathbb{R}^{ C \times N \times T / 2}, X_{ {even }} \in \mathbb{R}^{ C \times N \times T / 2}$. The 1D convolution operation in the interactive learning structure can be represented as $Conv{1}, Conv{2}, Conv{3}, Conv{4}$. The output after the first interactive learning is $X_{ {even }}^{\prime} \in \mathbb{R}^{ C \times N \times T / 2}, X_{ {odd }}^{\prime} \in \mathbb{R}^{ C \times N \times T / 2}$. $X_{ {even }}^{\prime}, X_{ {odd }}^{\prime}$ are subjected to one more interactive learning and the final subsequence are $X_{o d d_{-}  {out }}^{\prime} \in \mathbb{R}^{ C \times N \times T / 2}, X_{even_{-}  {out }}^{\prime} \in \mathbb{R}^{ C \times N \times T / 2}$ output. The operations in the interactive learning structure can be defined as:
\begin{equation}
X_{ {even }}, X_{o d d}={Split}(X)
\end{equation}
\begin{equation}
X_{o d d}^{\prime}={tanh}\left(D G C N\left(Conv{1}\left(X_{even}\right)\right)\right) \odot X_{ {odd }}
\end{equation}
\begin{equation}
X_{ {even }}^{\prime}={tanh} \left(D G C N\left(Conv{2}\left(X_{o d d}\right)\right)\right) \odot X_{ {even }}
\end{equation}
\begin{equation}
X_{o d d_{-}  {out }}=X_{ {odd }}^{\prime}+{tanh} \left(D G C N\left(Conv{3}\left(X_{ {even }}^{\prime}\right)\right)\right)
\end{equation}
\begin{equation}
X_{even_{-}{out }}=X_{ {even }}^{\prime}+{tanh} \left(D G C N\left(Conv{4}\left(X_{ {odd }}^{\prime}\right)\right)\right)
\end{equation}
where $\odot$ is denoted Hadamard product, $tanh$ is the activation function, and $DGCN$ denotes the dynamic graph convolution in the DGCN module. This interactive learning strategy allows the spatial-temporal features of each other to be captured between subsequences.

\subsection{Dynamic Graph Convolution}

The structure of the DGCN module is shown in Figure 2c, which consists of two main modules: graph generator and fusion graph convolution. DGCN module utilizes its generated graph structure to explore deeper spatial correlations better and increase the model's performance for capturing spatial heterogeneity. As shown in Figure 2c, the graph generator takes the hidden features $H\in\mathbb{R}^{ C\times N\times T}$ as input and the pre-defined initial adjacency matrix $A\in\mathbb{R}^{N\times N}$ as the first input. After the input $H\in\mathbb{R}^{ C\times N\times T}, A\in\mathbb{R}^{N\times N}$ is processed by diffusion graph convolution \cite{DCRNN}, it is then fed to the MLP to generate the matrix $A^{\prime}\in\mathbb{R}^{N\times N}$ with spatial-temporal feature:
\begin{equation}
A^{\prime} = SoftMax(MLP(GCN(H, A)))
\end{equation}
where $MLP$ denotes multilayer perceptron, and $GCN$ denotes diffusion graph convolution operation. 

The generated matrix $A^{\prime}\in\mathbb{R}^{N\times N}$ is discrete. Since this variable needs to be sampled during the training process to ensure the sampling process to be derivable, here we use Gumbel reparameterization \cite{gumbel-softmax} :
\begin{equation}
\begin{aligned}
A_{{learn}}&=GumbelSoftmax(A^{\prime}) \\
&=SoftMax(\left(\log \left(A^{\prime}\right)-\log(-\log(\mathbf{g}))\right) / \tau)
\end{aligned}
\end{equation}
where $\mathbf{g} \sim Gumbel(0,1)$ is a random variable. $\tau$ is the softmax temperature and is set to 0.5. $A_{learn}\in\mathbb{R}^{N\times N}$ is the adjacency matrix generated by the graph generator, and it can simulate the dynamic correlations generated between graph nodes.

In addition to the adjacency matrix $A_{learn}$ generated by the graph generator, we define an adaptive adjacency matrix\cite{GraphWaveNet} $A_{apt}\in\mathbb{R}^{N\times N}$ which can be expressed as :
\begin{equation}
A_{a p t}={SoftMax}\left(Relu\left(E_{1} E_{2}^{\top}\right)\right)
\end{equation}
where $E_{1} \in \mathbb{R}^{N \times c}$ and $E_{2} \in \mathbb{R}^{N \times c}$ are learnable parameters. The initial value of $A_{apt}$ is the adjacency matrix $A\in\mathbb{R}^{N\times N}$ pre-defined by a priori graph. 

As shown in Figure 2c, we fuse $A_{apt}$ with $A_{learn}$ using the adaptive fusion structure. After fusion, the resulting adjacency matrix $A_{dyn}$ is fed into the diffusion GCN for dynamic correlations simulation between nodes and and exploring the unseen node connections in the road network. This fusion operation can be defined as follows:
\begin{equation}
A_{d y n}=\alpha A_{a p t}+(1-\alpha) A_{l e a r n}
\end{equation}
where $A_{dyn}$ is the dynamic adjacency matrix after fusion and $\alpha$ is the learnable adaptive parameter factor.

In STIDGCN, we use diffusion GCN in three modules: graph generator, fusion graph convolution and concatenation fusion respectively. We define the input of diffusion GCN as $X_{in}\in\mathbb{R}^{C\times N\times T}$ uniformly. 

In the graph generator, the diffusion GCN used is defined as:
\begin{equation}
G C N\left(X_{in}, A_{a p t} \right)=\sum_{k=0}^{K} A_{a p t}^{k} X_{in} W
\end{equation}
where $A_{apt}\in\mathbb{R}^{N\times N}$ is the adaptive adjacency matrix, $k$ is the diffusion step, $K$ is the maximum number of diffusion step, and $W\in\mathbb{R}^{N\times N}$ denotes the parameter matrix. 

In the fusion graph convolution module, $A_{dyn}$ is the adjacency matrix of the input to the diffusion GCN, where the diffusion GCN can be defined as:	
\begin{equation}
G C N\left(X_{in}, A_{d y n}\right)=\sum_{k=0}^{K} A_{d y n}^{k} X_{in} W
\end{equation}

\begin{table*}[]
\caption{Dataset Description.}
\centering
\resizebox{15cm}{!}{
\begin{tabular}{cccccc}
\toprule
Datasets & Nodes & Edges & TimeSteps & Time   Range              & Missing Rate \\ \hline
PEMS03   & 358   & 547   & 26208     & 09/01/2018   - 11/30/2018 & 0.672\%             \\
PEMS04   & 307   & 340   & 16992     & 01/01/2018   - 02/28/2018 & 3.182\%             \\
PEMS07   & 883   & 866   & 28224     & 05/01/2017   - 08/31/2017 & 0.452\%             \\
PEMS08   & 170   & 295   & 17856     & 07/01/2016   - 08/31/2016 & 0.696\%             \\ \toprule
\end{tabular}
}
\end{table*}

As shown in Figure 2a, the spatial-temporal features obtained from the interactive learning structure are fed to a diffusion GCN module after recombining them in a temporally indexed order. Here the diffusion GCN module is used for feature capture and correction of the whole sequence data. Unlike before, we use two matrices in this diffusion GCN module, a pre-defined initial adjacency matrix $A\in\mathbb{R}^{N\times N}$ and a dynamic adjacency matrix $A_{dyn}\in\mathbb{R}^{N\times N}$ generated by the interactive learner structure. For the initial adjacency matrix $A\in\mathbb{R}^{N\times N}$, we use directed graphs, and the forward and backward transition matrices of $A$ are denoted as $A_{f}=A / {rowsum}(A)$ and $A_{b}=A^{T} / {rowsum}\left(A^{T}\right)$, respectively. The diffusion GCN module in this context can be defined as:
\begin{equation}
\begin{aligned}
&GCN(X_{in}, A, A_{d y n})
\\&=\sum_{k=0}^{K} (A_{f}^{k} X_{in} W_{1}+A_{b}^{k} X_{in} W_{2}+A_{d y n}^{k} X_{in} W_{3})
\end{aligned}
\end{equation}

The DGCN module can explore the unseen node connections in the road network to capture hidden spatial correlations. DGCN module can also dynamically simulate the dynamic associations generated between nodes over time based on the input traffic data. The DGCN module is embedded in the interactive learning structure to enhance the capture of temporal correlation using the captured spatial information during the training process.
\section{Experiment Results and Analysis}\label{Experiment Results and Analysis}

To evaluate the performance of STIDGCN, we conducted extensive experiments on four real-world freeway traffic flow datasets and verified the functionality of each STIDGCN module through ablation experiments.

\subsection{Datasets}
In our experiments, we used four real-world public available traffic flow datasets \cite{STSGCN}, which are PEMS03, PEMS04, PEMS07, and PEMS08. All four datasets were collected by the Caltrans Performance Measurement System (PeMS) \cite{PEMS} every 30 seconds in real-time. These data are eventually aggregated into 5-minute time observations, so there are 12 observations for an hour, and our goal is to predict the traffic data for the next hour. The dataset details are shown in Table 2. The graph's adjacency matrix in the experiments is constructed based on the distance between sensors in these real traffic road networks.

\subsection{Baseline Methods}
We compared STIDGCN with 15 baseline methods, which are as follows:
\begin{itemize}
    \item HA: Historical Average uses the average results of historical data to predict future data.
    \item VAR \cite{VectorAR}: Vector Auto-Regression is a time series model that captures the temporal correlation of traffic series.
    \item SVR \cite{SupportVR}: Support Vector Regression is a machine learning method that uses support vector machines to do regression on traffic sequences.
    \item LSTM \cite{LSTM/}: Long-Short Term Memory is a neural network-based model that can effectively capture time series correlation.
    \item TCN \cite{TCN}: Temporal Convolutional Neural Network is implemented by stacked causal dilation convolution to capture time series correlation efficiently.
    \item DCRNN \cite{DCRNN}: Diffusion Convolutional Recurrent Neural Network is an encoder-decoder structure that combines diffusion GCN with GRU to capture the spatial-temporal dependencies of traffic data.
    \item STGCN \cite{STGCN}: Spatial-Temporal Graph Convolutional Network combines the spectral GCN with 1D convolution to capture spatial-temporal dependencies.
    \item ASTGCN \cite{ASTGCN}: Attention based spatial-temporal graph convolutional network captures spatial-temporal dependencies by designing spatial and temporal attention mechanisms, respectively.
    \item Graph WaveNet \cite{GraphWaveNet}: Graph WaveNet combines gated TCN with spatial GCN and proposes an adaptive adjacency matrix to learn dynamic spatial correlations.
    \item AGCRN \cite{AGCRN}: Adaptive Graph Convolutional Recurrent Network is a model that combines GCN with GRU using an adaptive graph structure.
    \item STSGCN \cite{STSGCN}: Spatial-Temporal Synchronous Graph Convolutional Network is a GCN model that constructs multiple local spatial-temporal graphs to capture spatial-temporal dependencies synchronously.
    \item STFGNN \cite{STFGNN}: Spatial-Temporal Fusion Graph Neural Networks efficiently learn hidden correlations by performing fusion operations on the generated spatial-temporal graphs.
    \item ASTGNN \cite{ASTGNN}: Attention based Spatial-Temporal Graph Neural Network is a self-attention trafﬁc forecasting model that combines a time-trending self-attention mechanism with a dynamic GCN.
    \item SCINet \cite{SCINet}: Sample Convolution and Interaction Network uses an interactive convolutional structure, allowing multi-resolution processing of sequence data and expanding the receptive ﬁeld of the convolution operation.
    \item STG-NCDE \cite{STG-NCDE}: Spatial-Temporal Graph Neural Controlled Differential Equation uses spatial-temporal NCDEs to process traffic data and is a controlled differential equation method.
\end{itemize}

\begin{table*}[]
\caption{Comparison of STIDGCN and baselines on four trafﬁc datasets.}
\resizebox{\linewidth}{!}{
\begin{tabular}{c|ccc|ccc|ccc|ccc}
\toprule
\multirow{2}{*}{Methods} & \multicolumn{3}{c|}{PEMS03}                         & \multicolumn{3}{c|}{PEMS04}                         & \multicolumn{3}{c|}{PEMS07}                        & \multicolumn{3}{c}{PEMS08}                        \\ \cline{2-13}
                         & MAE            & RMSE           & MAPE             & MAE            & RMSE           & MAPE             & MAE            & RMSE           & MAPE            & MAE            & RMSE           & MAPE            \\ \hline
HA                       & 31.58          & 52.39          & 33.78\%          & 38.03          & 59.24          & 27.88\%          & 45.12          & 65.64          & 24.51\%         & 34.86          & 59.24          & 27.88\%         \\
VAR                      & 23.65          & 38.26          & 24.51\%          & 24.54          & 38.61          & 17.24\%          & 50.22          & 75.63          & 32.22\%         & 19.19          & 29.81          & 13.10\%         \\
SVR                      & 21.97          & 35.29          & 21.51\%          & 28.70          & 44.56          & 19.20\%          & 32.49          & 50.22          & 14.26\%         & 23.25          & 36.16          & 14.64\%         \\
LSTM                     & 21.33          & 35.11          & 23.33\%          & 26.77          & 40.65          & 18.23\%          & 29.98          & 45.94          & 13.20\%         & 23.09          & 35.17          & 14.99\%         \\
TCN                      & 19.32          & 33.55          & 19.93\%          & 23.22          & 37.26          & 15.59\%          & 32.72          & 42.23          & 14.26\%         & 22.72          & 35.79          & 14.03\%         \\
DCRNN                    & 17.99          & 30.31          & 18.34\%          & 21.22          & 33.44          & 14.17\%          & 25.22          & 38.61          & 11.82\%        & 16.82          & 26.36          & 10.92\%         \\
STGCN                    & 17.55          & 30.42          & 17.34\%          & 21.16          & 34.89          & 13.83\%          & 25.33          & 39.34          & 11.21\%         & 17.50          & 27.09          & 11.29\%         \\
ASTGCN                   & 17.34          & 29.56          & 17.21\%          & 22.93          & 35.22          & 16.56\%          & 24.01          & 37.87          & 10.73\%         & 18.25          & 28.06          & 11.64\%         \\
GWN                      & 14.79          & 25.51          & \textbf{14.32}\% & 19.36          & 31.72          & 13.31\%          & 21.22          & 34.12          & 9.07\%          & 15.07          & 23.85     & 9.51\%          \\
STSGCN                   & 17.48          & 29.21          & 16.78\%          & 21.19          & 33.65          & 13.90\%          & 24.26          & 39.03          & 10.21\%         & 17.13          & 26.80          & 10.96\%         \\
AGCRN                    & 15.98          & 28.25          & 15.23\%          & 19.83          & 32.26          & 12.97\%          & 22.37          & 36.55          & 9.12\%          & 15.95          & 25.22          & 10.09\%         \\
STFGNN             & 16.77          & 28.34          & 16.30\%          & 19.83          & 31.88          & 13.02\%          & 22.07          & 35.80          & 9.21\%          & 16.64          & 26.22          & 10.60\%         \\
ASTGNN                   & 14.78 & 25.00 & 14.79\%          & 18.60 & 30.91 & 12.36\% & 20.62          & 34.00          & 8.86\%          & 15.00 & 24.70          &  9.50\% \\
SCINet             & 15.25          & 24.58          & 14.54\%          & 19.30          & 31.28          & \textbf{12.05\%}          & 21.56          & 34.37          & 9.03\%          & 15.76          & 24.65          & 10.01\%         \\
STG-NCDE                 & 15.57          & 27.09          & 15.06\%          & 19.21          & 31.09          & 12.76\%          & 20.53 & 33.84 & 8.80\% & 15.45          & 24.81          & 9.92\%          \\ \hline
STIDGCN                  & \textbf{14.55} & \textbf{24.42} & 14.68\% & \textbf{18.42} & \textbf{29.81} & 12.27\% & \textbf{19.28} & \textbf{32.26} & \textbf{8.36\%} & \textbf{14.03} & \textbf{23.35} & \textbf{9.15\%} \\
\toprule
\end{tabular}
}
\end{table*}

\subsection{Settings}
All datasets used in the experiments are divided into the training set, validation set, and test set according to the ratio of 6:2:2. Z-score normalization is applied to standardize the datasets before feeding them to the network. For the missing traffic data, we mask them, i.e., we do not consider these missing data (value 0). We use data from 12 continuous-time steps in the past one hour to predict data from 12 continuous future time steps of the next hour. 

Experiments are conducted under a computer environment with one Intel(R) Xeon(R) Gold 6230 CPU @ 2.10GHz and one NVIDIA Tesla V100 GPU card. We train STIDGCN using the Ranger optimizer \cite{Ranger21} with the initial learning rate set to 0.001. The batch size is set to 64, and the training epoch is 500, where we set the early stop mechanism. The model converges to an end at around epoch 300.

We selected three standard metrics to evaluate all methods' performance, namely mean absolute error (MAE), mean absolute error (MAPE), and root mean square error (RMSE), which are defined as follows:
\begin{equation}
M A E=\frac{1}{N} \sum_{i=1}^{N}\left|Y_{i}-\hat{Y}_{i}\right|
\end{equation}
\begin{equation}
M A P E=\frac{100 \%}{N} \sum_{i=1}^{N}\left|\frac{Y_{i}-\hat{Y}_{i}}{Y_{i}}\right|
\end{equation}
\begin{equation}
R M S E=\sqrt{\frac{1}{N} \sum_{i=1}^{N}\left(Y_{i}-\hat{Y}_{i}\right)^{2}}
\end{equation}
where $N$ denotes the number of samples, $Y_{i}$ denotes the ground truth, and $\hat{Y}_{i}$ denotes the prediction value.

\subsection{Comparison and Analysis of Results}

Table 3 shows the forecasting results of STIDGCN and the comparison model for the next hour (12 timesteps) on four test sets of real traffic flow data. Our proposed STIDGCN performs better than all baseline methods on these four traffic datasets, except for the MAPE metric in PEMS03 and PEMS04, which is slightly worse than Graph WaveNet and SCINet, respectively. STIDGCN on PEMS07 and PEMS08, the results have more noticeable improvement. In PEMS07, STIDGCN improves the state-of-the-art method by 6.1\%, 4.7\%, and 5.0\% in MAE, RMSE and MAPE, respectively. In PEMS08, STIDGCN improves the state-of-the-art method by 6.5\%, 2.1\%, and 3.7\% in MAE, RMSE and MAPE, respectively.

\begin{figure*}[htbp]  
\centering
\includegraphics[width=17cm]{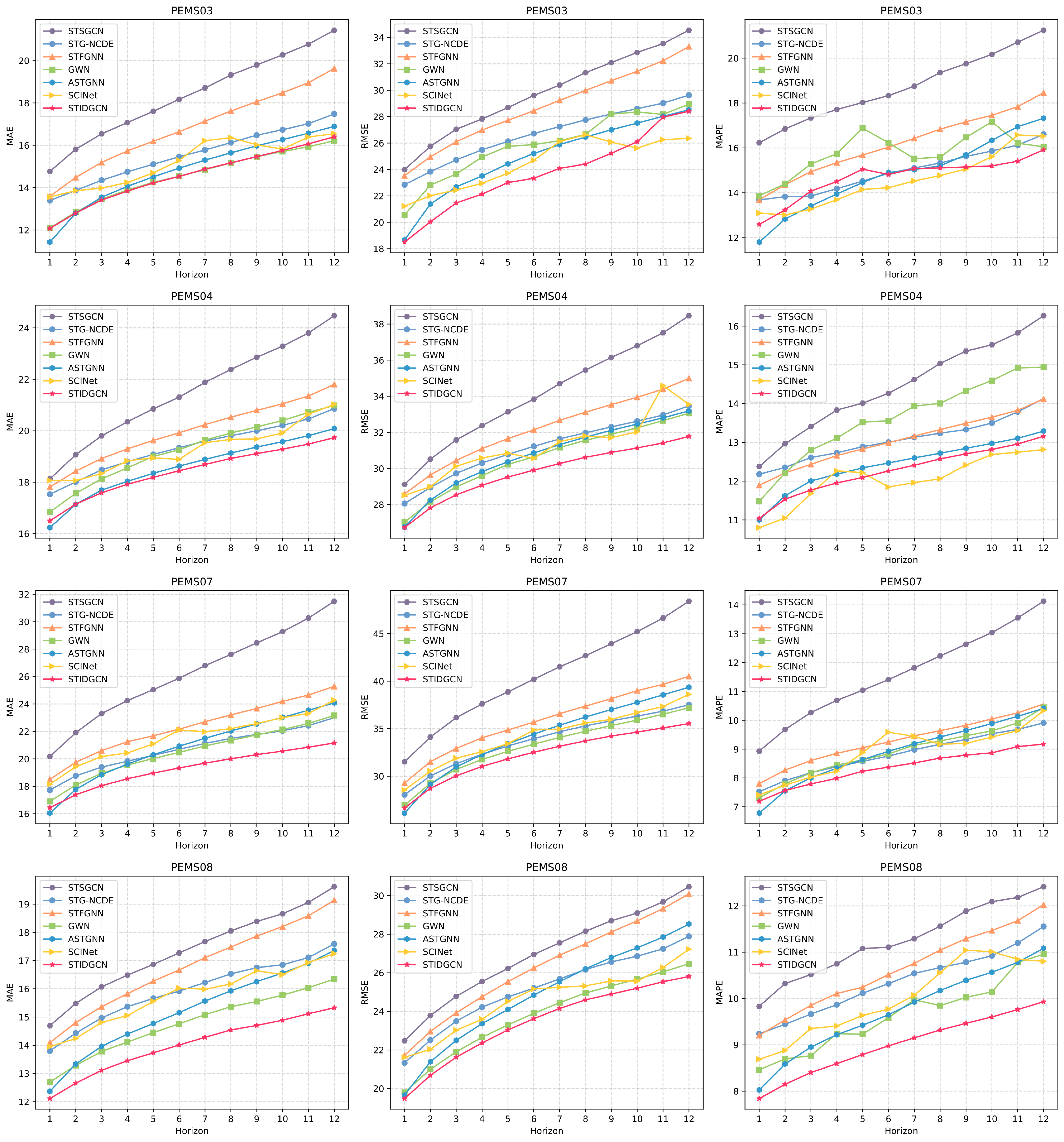}
\caption{Forecasting performance comparison at each horizon on four datasets. STIDGCN performs best in PEMS07, PEMS08, and PEMS04. STIDGCN outperforms ASTGNN at time steps 10, 11, and 12. STIDGCN has a better long-range forecast ability.}  
\label{fig4}   
\end{figure*}

The results in Table 3 show that statistical methods (HA, VAR), traditional machine learning methods (SVR), LSTM and TCN perform poorly because these models only consider temporal correlation and neglect the complex spatial correlation in traffic data. Spatial-temporal GCN models represented by STGCN and DCRNN perform better because both temporal and spatial correlations are considered. In addition, models based on attention mechanisms (ASTGCN, ASTGNN) also perform well because attention mechanisms can capture the temporal correlation of long sequences. Graph WaveNet performs even better than some recently proposed models (STFGNN, STG-NCDE). Graph WaveNet is the embedding of diffusion GCN into TCN, which has an excellent ability to capture temporal correlation. This compact dependency capture is close to the synchronous capture of spatial-temporal features. Although STSGCN uses a synchronous method to capture spatial-temporal features, it uses a simple sliding window to capture temporal correlations. It indicates that STSGCN downplays the capture of temporal correlation, and its overall performance is not very good even when the spatial correlation is captured effectively. It is worth noting that SCINet achieves good performance even without considering spatial correlation. It also demonstrates the effectiveness of interactive learning and the importance of spatial modeling for traffic data. NCDE as a new deep learning model, even with the good results obtained, its performance is not as good as the proposed STIDGCN due to its capturing spatial-temporal features in series. Our proposed STIDGCN uses an interactive learning strategy for synchronous capturing of correlations. Its DGCN module can explore the unseen node connections in the road network to capture hidden spatial correlations and simulate the generation of dynamic associations between nodes over time.

Figure 4 shows the variation of MAE, RMSE, and MAPE for a portion of the model on four traffic datasets with increasing forecasting time horizons. As the forecast horizon increases, the forecast difficulty will also change; MAE, RMSE, and MAPE will continue to increase. Our proposed STIDGCN uses an interactive learning strategy where sequences learn each other's spatial-temporal features to obtain long-range forecasting capabilities. As shown in Figure 4, STIDGCN performs better than ASTGNN based on the attention mechanism, even in long-range forecasting.

\subsection{Ablation Study}

\begin{figure*}[htbp]  
\centering
\includegraphics[width=15cm]{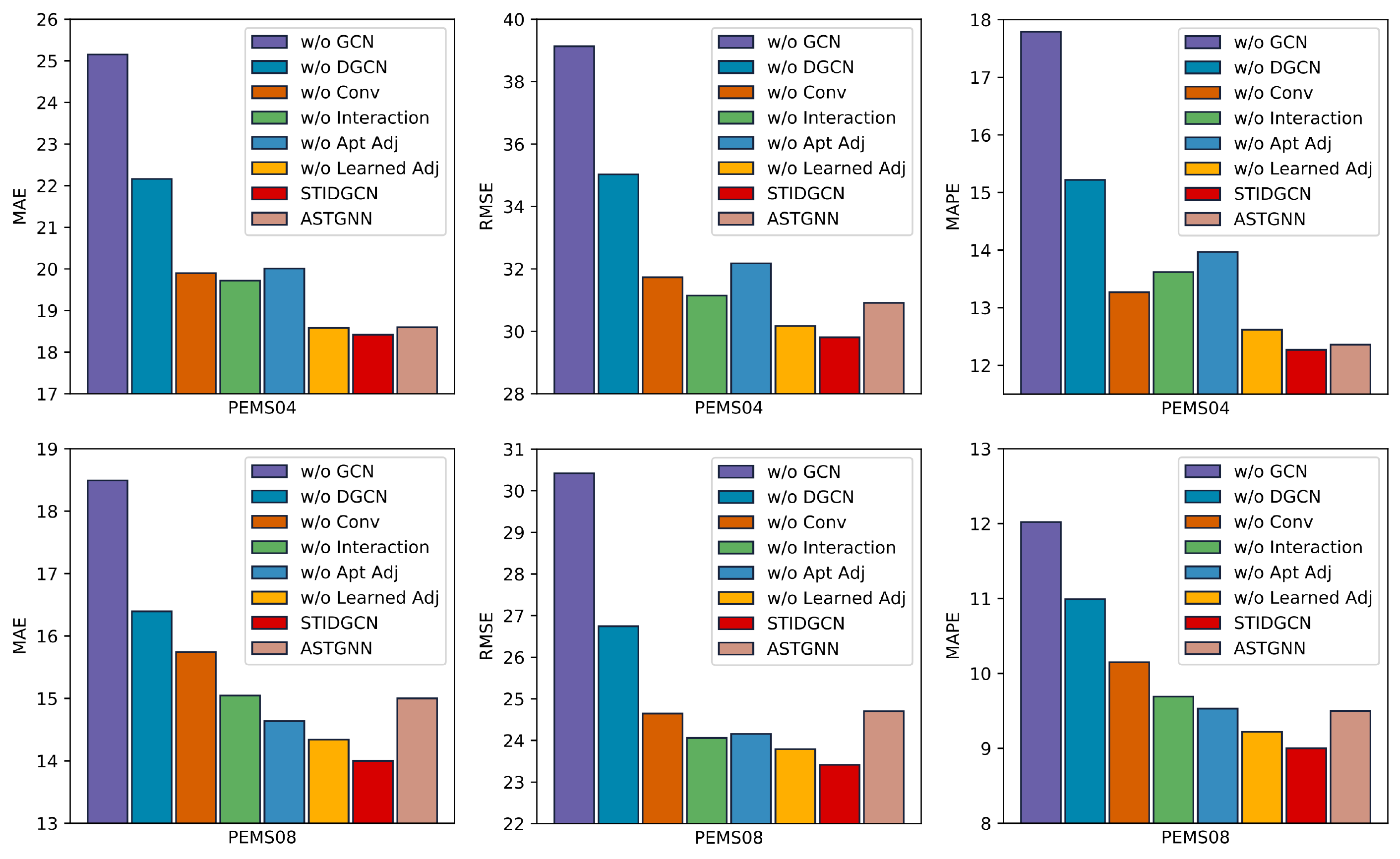}
\caption{Component analysis of STIDGCN. Each of the components in STIDGCN contributes a positive effect. Among them, DGCN and the interactive learning structure are the components that influence the model most.} 
\label{fig5}   
\end{figure*}

To further evaluate which components of STIDGCN are the key components affecting the model results, we conducted ablation experiments on PEMS04 and PEMS08 datasets. We designed six STIDGCN variants, which are as follows:
\begin{itemize}
\item w/o GCN: On the base of STIDGCN, the diffusion GCN module is removed.
\item w/o DGCN: On the base of STIDGCN, the DGCN module is replaced with a normal diffusion GCN. The adjacency matrix input to the GCN is the pre-defined initial adjacency matrix.
\item w/o Conv: On the base of STIDGCN, remove the 1D convolution module from the interactive learning structure.
\item w/o Interaction: On the base of STIDGCN, the interactive learning structure is replaced by a TCN and connected in series with the dynamic convolution module. The TCN is set to 6 layers of convolution, and the number of feature channels is 64.
\item w/o Apt Adj: On the base of STIDGCN, the adaptive adjacency matrix of the DGCN module is removed. The adjacency matrix of the input graph generator is replaced with the pre-defined initial adjacency matrix.
\item w/o Learned Adj: On the base of STIDGCN, remove the graph generator structure, keep the adaptive adjacency matrix, and change the fusion GCN to diffusion GCN in the DGCN module.
\end{itemize}

\begin{figure*}[htbp]  
\centering
\includegraphics[width=15cm]{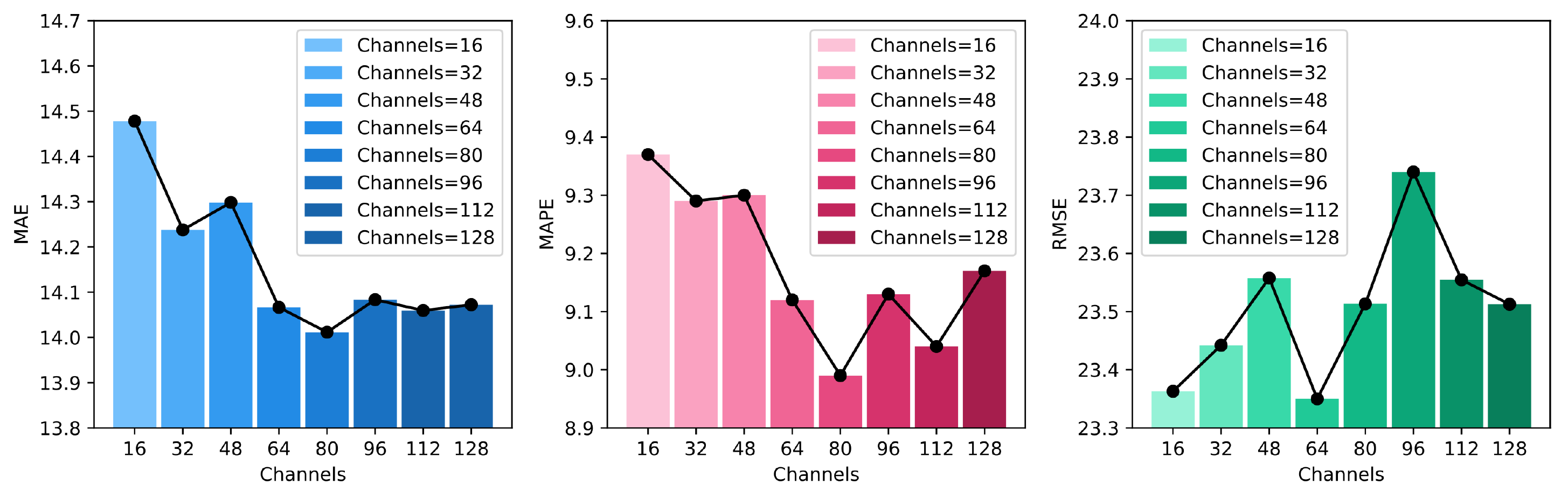}
\caption{Effect of the number of feature channels on model performance. Increasing the number of feature channels from 16 to 128 shows that STIDGCN has the best performance in the 64-80 interval.} 
\label{fig6}   
\end{figure*}

The results of the ablation experiments are shown in Figure 5. The effects of these components on the model as a whole have essentially similar distributions on the PEMS04 and PEMS08 datasets. We also compared ASTGNN in ablation experiments. Firstly, GCN is crucial for STIDGCN, and secondly, our proposed IDGCN module is crucial for the overall model performance improvement. The 1D convolution used to expand the receptive field is a crucial component of the interactive learning structure, and ablation experiments show that it can significantly improve model performance. The model performance deteriorated after replacing the interactive learning structure with TCN, and ASTGNN‘s combined performance outperforms STIDGCN. It verifies that synchronous feature capture in the interactor is more effective than capture in series like TCN or ASTGNN. In addition, we conducted an ablation study of the two adjacency matrices defined inside the DGCN module, and Figure 5 shows that the adaptive adjacency matrix is vital to the model. The learnable adjacency matrix also plays a significant role and cooperates with the adaptive adjacency matrix to generate the dynamic adjacency matrix. As shown in Figure 5, ASTGNN‘s performance outperforms STIDGCN when STIDGCN removes the dynamic adjacency matrix. It shows the necessity of a dynamic adjacency matrix for graph convolution. This dynamic adjacency matrix allows graph convolution better to capture the hidden spatial correlations in the traffic data. Therefore, our proposed two core structures (interactive learning and dynamic graph convolution) are effective.
 
\subsection{Effect of Different Structure Conﬁgurations}
To further investigate the effects of hyperparameter settings and model structure settings, we conducted experiments on the PEMS08 dataset. STIDGCN does not take the form of repeatedly stacked modules to capture features, so the only hyperparameters related to the model structure are the number of feature channels. As shown in Figure 6, the number of feature channels specifically affects STIDGCN. STIDGCN's performance does not improve consistently as the number of feature channels increases. STIDGCN's performance also tends to level off when the number of feature channels is increased to a specific value. While feature channels range from 64 to 80, STIDGCN performs best. However, increasing the number of feature channels leads to more parameters and a longer inference time for the model. Therefore, the number of feature channels of STIDGCN is finally selected as 64.

In addition, we have made a few changes to the submodules in the interactive learning structure to demonstrate that the current structure is the most reasonable. The interactive learning structure consists of four separate convolutional modules and one weight-sharing DGCN module. The convolution module is designed to extract correlations between different time steps in the current sequence and broadens the model's receptive field in the interactive learning architecture. Meanwhile, the weight-sharing DGCN module can interactively learn spatial correlations in different sequences. We have made the following changes to the IDGCN module:

\begin{itemize}
\item \textit{IDGCN Conv×1}: Replace the 4 separate convolutional modules with a single weight-sharing convolutional module in the interactive learning structure.
\item \textit{IDGCN DGCN×2}: The weight-sharing DGCN module in the interaction learning structure is replaced by 2 separate DGCN modules. Two separate DGCN modules are used to capture the spatial correlation of the sequences during the two interactive learning sessions, respectively.
\item \textit{IDGCN DGCN×4}: The weight-sharing DGCN modules in the interaction learning structure are replaced by 4 separate DGCN modules. The 4 DGCN modules capture the spatial correlation of the sequences during the two interactive learning processes, respectively.
\item \textit{IDGCN Interaction×1}: The number of interactive learning in the interactive learning structure is changed from 2 to 1, i.e., two convolution operations and one graph convolution operation are removed.
\item \textit{IDGCN Interaction×4}: The number of interactive learning in the interaction learning structure is changed from 2 to 4, i.e., the interactive learning operation in the IDGCN module is repeated twice.
\end{itemize}

As shown in Table 4, the changed structure affects STIDGCN. The results in Table 4 show that the weight-sharing convolution module is not more effective than the separate convolution module. DGCN module has the opposite performance; adding the DGCN module does not lead to better performance. The weight-sharing DGCN module is equivalent to stacking DGCN modules in interactive learning, allowing STIDGCN to have a larger receptive field in spatial terms and nodes to make associations with more distant nodes. A separate DGCN module does not give a shared receptive field and can lead to poorer model performance. STIDGCN uses a synchronous strategy to capture spatial-temporal dependencies in the interactive learning structure. Therefore, reducing the number of interactive learning means reducing the capture of spatial-temporal dependencies, and this operation will undoubtedly lead to poorer performance of STIDGCN. Experiments show that two interactive learning is sufficient to fit the features, and increasing the interactive learning operation in the IDGCN module leads to a poorer ability to fit the features.

\begin{table}[]
\caption{IDGCN module configuration analysis.}
\resizebox{\linewidth}{!}{
\centering
\begin{tabular}{cccc}
\toprule
Structure                    & MAE   & RMSE  & MAPE \\ \hline
\textit{IDGCN Conv×1}        & 14.33 & 23.57 & 9.22\% \\
\textit{IDGCN DGCN×2}        & 14.60 & 23.87 & 9.69\% \\
\textit{IDGCN DGCN×4}        & 14.64 & 24.06 & 9.91\% \\
\textit{IDGCN Interaction×1} & 14.32 & 23.74 & 9.32\% \\
\textit{IDGCN Interaction×2} & 14.81 & 23.87 & 9.81\% \\
STIDGCN                      & \textbf{14.03}   & \textbf{23.35}   & \textbf{9.15\%}  \\ \toprule
\end{tabular}
}
\end{table}

\subsection{Computation Time}

In this part, we compare the computation cost of STIDGCN with part of the baseline models on the PEMS08 dataset in Table 5. STIDGCN does not suffer from severe computation costs while having outstanding performance. Compared with other state-of-the-art baselines like ASTGNN and STG-NCDE, STIDGCN achieves better performance while significantly reducing computation costs. Although ASTGNN has emergent performance, it processes data in an autoregressive manner, leading to high computation costs. Graph Wave has less computation cost due to parallel data processing and lightweight model. Although SCINet has the lowest computation cost, it has its deficiency in capturing spatial correlations. The computation cost of STIDGCN mainly comes from the dynamic graph convolution structure in its interactive learning. STIDGCN parallelizes data processing in a non-autoregressive manner to make the model more efficient.

\begin{table}[]
\caption{The computation cost on the PEMS08 dataset.}
\resizebox{\linewidth}{!}{
\centering
\begin{tabular}{cccc}
\toprule
\multirow{2}{*}{Model} & \multicolumn{2}{c}{Computation Time} \\ 
                       & Training (s/epoch)  & Inference (s)  \\ \hline
GWN                    & 10.64             & 1.20         \\
STSGCN                 & 61.28             & 12.40        \\
STFGNN                 & 21.93            & 4.45         \\
SCINet                 & 5.09              & 2.86         \\
ASTGNN                 & 44.48               & 23.70        \\
STG-NCDE               & 84.47            & 9.17        \\
STIDGCN                & 22.16             & 2.43        
\\ \toprule
\end{tabular}
}
\end{table}

\section{Conclusion}\label{Conclusion}
In this paper, we propose a dynamic graph generation model STIDGCN using an interactive learning strategy. Specifically, we embed dynamic GCN modules into the interactive learning structure to capture spatial-temporal dependencies synchronously. The application scenario of interactive learning strategy is time series forecasting that exhibits certain periodicity and trend. We propose a DGCN module to simulate dynamic spatial correlations, i.e., to use the input spatial-temporal information to generate a dynamic graph structure and cooperate with a pre-defined initial adjacency matrix. The DGCN module explores the unseen node connections in the road network to capture hidden spatial correlations and simulates the dynamic correlation generated between nodes over time. Experiments on four real datasets show that simultaneous learning and dynamic graph generation are essential for spatial-temporal forecasting. Our proposed model outperforms the state-of-the-art baseline. In interactive learning, feature learning is performed in uniform distribution among individual sequences. However, the degree of correlations between data is not always uniformly distributed. Therefore, we consider using attention mechanisms in interactive learning as a good improvement solution. 



\ifCLASSOPTIONcaptionsoff
  \newpage
\fi

\bibliographystyle{IEEEtran}
\bibliography{reference}

\end{document}